\documentclass[10pt]{article}

\usepackage[margin=1in]{geometry}
\usepackage{times}
\usepackage{microtype}
\usepackage[hidelinks]{hyperref}
\usepackage{url}

\usepackage{amsmath}
\usepackage{amsfonts}
\usepackage{amssymb}

\usepackage{graphicx}
\usepackage{booktabs}
\usepackage{tabularx}
\usepackage{adjustbox}
\usepackage{fancyvrb}

\usepackage{listings}
\usepackage{xcolor}

\lstdefinestyle{mypython}{
    language=Python,
    backgroundcolor=\color{white},
    commentstyle=\color{green!50!black},
    keywordstyle=\color{blue!60!black},
    stringstyle=\color{orange!60!black},
    basicstyle=\ttfamily\small,
    breaklines=true,
    showstringspaces=false,
    tabsize=2,
    frame=single,
    numbers=left,
    numberstyle=\tiny\color{black!50}
}

\usepackage{titlesec}

\titlespacing*{\section}
  {0pt}        % left
  {2.0ex}      % space before
  {1.2ex}      % space after

\titlespacing*{\subsection}
  {0pt}
  {1.6ex}
  {0.8ex}

\titlespacing*{\subsubsection}
  {0pt}
  {1.2ex}
  {0.6ex}

\usepackage[sorting=none, style=nature]{biblatex} 
\title{Gradient Boosted Risk Scores}

\author{
Costa Georgantas\textsuperscript{\textdagger,*},
Jonas Richiardi\textsuperscript{\textdagger}\\
\textsuperscript{\textdagger}Department of Radiology, Lausanne University Hospital and University of Lausanne, Switzerland\\
\textsuperscript{*}Corresponding author: \href{mailto:costa.georgantas@chuv.ch}{costa.georgantas@chuv.ch}
}

\date{}

\begin{document}

\maketitle

\section{Abstract}
Risk scores are an interpretable and actionable class of machine learning models with applications in medicine, insurance, and risk management. Unlike most computational methods, risk scores are designed to be computed by a human by attributing points to a data sample based on a limited set of criteria. The most common approaches for generating risk scores use linear regressions to estimate the effect of selected variables. We propose a simple and effective approach towards building compact and predictive risk scores. We provide an algorithm based on gradient boosting that is capable of modeling nonlinear effects, along with a C++ implementation with Python and R bindings. Through extensive empirical evaluation on twelve tabular datasets spanning regression, classification, and time-to-event tasks, we show that our method achieves competitive predictive performance while producing substantially more compact scores than regression-based alternatives, with 60\% fewer rules for classification tasks and 16\% fewer rules for time-to-event tasks on average, compared to AutoScore.

\section{Introduction}

Risk score models are rule-based algorithms that consist of allocating and summing up points from a reference table to produce a prediction. Unlike most machine learning techniques, risk scores are transparent, meaning that a human can directly assess the impact of each input variable on the model prediction. Risk scores also enable decision-makers to think strategically about risk by visually estimating the effect of varying input variables. These properties make risk scores particularly advantageous for problems that require a complete understanding of the model output. They have applications in credit risk \cite{thomas_survey_2000, pehlivanli_introducing_2024}, criminal recidivism \cite{noauthor_kentucky_nodate, kovalchuk_scoring_2022}, and healthcare \cite{antman_timi_2000}.

Risk scores are particularly prevalent in clinical settings, where they can be used for the stratification of patients for various diseases \cite{london_artificial_2019}. Many clinical risk scores (CRS) have been developed to estimate the risk of hospitalization, death, or other future adverse events based on clinical variables and test results \cite{dagostino_general_2008,smith_early_2014, visseren_2021_2021}. Due to their transparency and portability, scores have also been used as inclusion criteria for clinical trials. Score cutoffs allow for individuals with the same diagnosis but with different characteristics to be selected for treatment based on their risk of future adverse events. This is particularly important for applications where prevention is effective and for diseases that can arise from multiple etiologies, such as heart failure.

Current techniques for producing risk scores rely on using weights from regression algorithms to estimate variable effects. Typically, this is done with linear, logistic, or Cox regression for continuous, binary, and survival outcomes, respectively. However, this approach is problematic for two reasons. First, the resulting risk scores are constrained by the assumptions of the underlying regression model, which may not hold in practice. Second, the regression coefficients must be binned to create interpretable point-based scores, which introduces information loss and reduces predictive accuracy. A cutting plane algorithm for computing scores has also been proposed \cite{ustun_learning_2019}, although it requires extensive parameter tuning. AutoScore \cite{xie_autoscore_2020}, a machine learning–based framework, has been developed to address these limitations by automating the generation of interpretable risk scores while maintaining predictive performance. AutoScore relies on converting continuous variables into categorical ones before regressing to obtain coefficients, which still introduces a loss of information.

We propose to directly model risk scores as a set of decision rules. Similarly to gradient boosted trees methods, we sequentially aggregate and combine decision stumps to produce predictive risk scores in a single optimization step without post-processing. We demonstrate that our approach is superior to other interpretable methods on average and exhibits minimal performance loss in many cases compared to other non-transparent gradient-boosted techniques. We use a minimal set of hyperparameters and do not rely on tuning regularization coefficients. Our proposed method applies to continuous, binary, and survival objectives. We provide an efficient C++ implementation of this algorithm, along with Python and R bindings that enable scores to be generated with just a few lines of code.

To the best of our knowledge, GBRS is the first gradient-boosting–based approach explicitly designed to produce human-computable, point-based risk scores. Unlike rule-based methods such as RuleFit \cite{friedman_predictive_2008}, which generates rules from trees of varying depth and subsequently prune them using a sparse linear model, we restrict base learners to decision stumps with pre-specified or data-driven threshold cutoffs and directly aggregate repeated stump conditions into a compact additive score. This design yields transparent scoring tables that remain predictive while being straightforward to evaluate by hand. Similarly, while explainable boosting machines (EBMs) \cite{nori_interpretml_2019} learn additive shape functions that require computational evaluation, GBRS outputs explicit point contributions whose sum defines the prediction, so the explanation is the model itself; this is a fundamental difference between explainable and transparent models.

\section{Methods}

\subsection{Risk scores as additive point systems}

Gradient boosted risk scores (GBRS) is a method for constructing interpretable, human-computable risk scores using decision stumps learned via a gradient boosting procedure \cite{buhlmann_boosting_2007}. A risk score is an additive model in which a prediction is obtained by summing a small number of points associated with simple threshold-based rules on individual variables. Our central design goal is to learn such point systems directly, rather than approximating them post hoc from a black-box model.

Let $\mathcal{D} = \{(x_i, y_i)\}_{i=1}^N$ denote a dataset of $N$ samples, where each sample $i$ has $d$ covariates $x_i = (x_{i1}, \dots, x_{id}) \in \mathbb{R}^d$ and an outcome $y_i \in \mathbb{Y}$. For regression tasks, $\mathbb{Y}=\mathbb{R}$; for binary classification, $\mathbb{Y}=\{0,1\}$; and for survival analysis, $\mathbb{Y}$ consists of observed times $T_i$ and event indicators $E_i$.

GBRS models predictions as an additive score
\begin{equation}
F(x_i) = \beta_0 + \sum_{k=1}^{K} s_k(x_{ij_k}),
\label{eq:additive_score}
\end{equation}
where each $s_k(\cdot)$ is a univariate, piecewise-constant scoring function applied to a single covariate $j_k$. Each term contributes a fixed number of points depending on whether the covariate crosses a threshold, and the final prediction is obtained by summing all point contributions.

Without loss of generality, a univariate scoring function can be written as
\begin{equation}
s(x_{ij}) =
\begin{cases}
a_0, & x_{ij} < b_1, \\
a_1, & b_1 \le x_{ij} < b_2, \\
\vdots \\
a_K, & x_{ij} \ge b_K,
\end{cases}
\label{eq:piecewise_score}
\end{equation}
where $\{b_1,\dots,b_K\}$ are fixed cutoffs. This function can be rewritten as a sum of weighted indicator functions:
\begin{equation}
s(x_{ij}) = a_0 + \sum_{k=1}^{K} (a_k - a_{k-1})\, \mathbf{1}_{\{x_{ij} \ge b_k\}}.
\label{eq:indicator_form}
\end{equation}
Each indicator term corresponds exactly to a decision stump with a single threshold. Consequently, any univariate risk score lies in the linear span of decision stumps, motivating the use of additive stump models as a natural optimization framework.

\subsubsection{Gradient boosting with decision stumps}

GBRS learns the scoring function in Eq.~\eqref{eq:additive_score} using gradient boosting, restricting all base learners to decision stumps. At boosting iteration $m$, the model prediction is
\begin{equation}
\hat{y}_i^{(m)} = \sum_{t=1}^{m} f_t(x_i),
\end{equation}
where each $f_t$ belongs to the space of decision stumps
\[
\mathcal{F}_{\text{stump}} =
\left\{
f(x_i) =
\begin{cases}
\gamma_0, & x_{ij} < \tau_j, \\
\gamma_1, & x_{ij} \ge \tau_j,
\end{cases}
\;\middle|\;
j \in \{1,\dots,d\},\; \tau_j \in \mathcal{T}_j
\right\}.
\]
Here, $\mathcal{T}_j$ is a finite set of candidate thresholds for feature $j$, specified either by the user (e.g., clinically meaningful cutoffs) or derived from training-set quantiles.

At each iteration $m$, GBRS computes pseudo-residuals as the negative gradient of a task-specific loss function:
\begin{equation}
g_i^{(m)} = -\left.
\frac{\partial L(y_i, \hat{y}_i)}{\partial \hat{y}_i}
\right|_{\hat{y}_i = \hat{y}_i^{(m-1)}}.
\end{equation}
A decision stump is then fit to these pseudo-residuals by selecting a feature $j$, threshold $\tau_j$, and leaf values $(\gamma_0,\gamma_1)$ that minimize the loss after an update. Predictions are updated as
\begin{equation}
\hat{y}_i^{(m)} = \hat{y}_i^{(m-1)} + \nu f_m(x_i),
\end{equation}
where $\nu \in (0,1]$ is a fixed learning rate.

\subsubsection{Closed-form leaf updates}

\paragraph{Continuous outcomes}
For squared-error regression, pseudo-residuals are $g_i^{(m)} = y_i - \hat{y}_i^{(m-1)}$. For a fixed split $(j,\tau)$, the optimal leaf values are the within-group means:
\begin{equation}
\gamma_g =
\frac{1}{|G_g|} \sum_{i \in G_g} g_i^{(m)},
\qquad
G_0 = \{i : x_{ij} < \tau_j\},\;
G_1 = \{i : x_{ij} \ge \tau_j\}.
\end{equation}

\paragraph{Binary outcomes}
For binary classification, GBRS models log-odds $f_i$ with probabilities $p_i=\sigma(f_i)$. Using a Newton step, leaf values are computed as
\begin{equation}
\gamma_g =
\frac{\sum_{i \in G_g} (y_i - p_i)}{\sum_{i \in G_g} p_i(1-p_i)},
\end{equation}
corresponding to a two-leaf logistic regression update.

\paragraph{Time-to-event outcomes}
For survival analysis, GBRS optimizes a pairwise ranking loss over comparable pairs $(i,j)$ with $E_i=1$ and $T_i < T_j$:
\begin{equation}
\mathcal{L}_{\text{rank}} =
\frac{1}{|\mathcal{P}|}
\sum_{(i,j)\in\mathcal{P}}
\log\!\left(1+\exp(f_j - f_i)\right).
\end{equation}
Pseudo-residuals are given by the negative gradient of this loss, and stump leaf values are computed as the mean pseudo-residual within each group $G_0,G_1$.

\subsubsection{Score aggregation and transparency}

Because decision stumps have no hierarchical dependencies, identical split conditions $(j,\tau_j)$ may be selected in multiple boosting iterations. GBRS aggregates such stumps by summing their coefficients, producing an equivalent but more compact representation. The resulting model is a sparse set of threshold-based rules, each contributing a fixed number of points to the total score. An example is shown in Table \ref{score_ex}.

In continuous tasks, points represent additive effects on the outcome. In binary tasks, the summed score is mapped through a logistic function to obtain a probability. In survival analysis, scores correspond to relative log-risk, and exponentiating the score yields hazard ratios. GBRS uses the number of boosting iterations and the learning rate as its primary hyperparameters, with optional subsampling and user-defined thresholds.

GBRS models predictions additively on a fixed scale: the identity scale for continuous outcomes, the log-odds scale for binary outcomes, and the log-risk (or log-hazard) scale for time-to-event outcomes. Each score coefficient therefore represents an additive contribution on this scale within a predefined threshold region. Because GBRS restricts base learners to decision stumps and does not model interactions, the resulting score is a generalized additive model with piecewise-constant components.

It is important to note that these coefficients should be interpreted as additive contributions to the model score, rather than as marginal or causal effects. As in other additive models learned by gradient boosting, coefficients are estimated sequentially and may be influenced by feature correlations. Nevertheless, the absence of interactions and the use of fixed thresholds ensure that each point contribution has a direct and transparent interpretation on the chosen scale, enabling meaningful comparison of risk across variable ranges.

\subsection{GBRS Implementation}
The GBRS library is installable as R and Python packages from a GitLab repository \url{https://gitlab.com/cgeo/GBRS}. The optimization algorithm is implemented in C++ using the Eigen header-only library \cite{gael_guennebaud_eigen_2010} and parallelized with OpenMP \cite{dagum_openmp_1998}. We provide declarative APIs that resemble those of other commonly used Python and R libraries, along with methods for printing score outputs, as showcased in Table \ref{score_ex}. Full examples of Python and R usage are available in the repository, and small code snippets are provided below:
\begin{lstlisting}[style=mypython]
Python:
gbrs_model = GBRS(n_iter=500, lr=0.05, n_quantiles=4)
gbrs_model.fit(X_train, y_train)
preds_gbrs = gbrs_model.predict(X_test)
gbrs_model.print()
\end{lstlisting}
\begin{lstlisting}[style=mypython]
R:
gbrs_model <- gbrs("y ~ x1 + x2 + x3", train_set)
preds_gbrs <- predict(gbrs_model, test_set)
print(gbrs_model)
\end{lstlisting}

\section{Results}

 GBRS performance was evaluated in continuous, binary, and time-to-event prediction tasks. We benchmark against traditional linear, logistic, and Cox regression, XGBoost \cite{chen_xgboost_2016}, which is still a SOTA method for tabular data prediction \cite{grinsztajn_why_2022}, and AutoScore \cite{xie_autoscore_2020}. Autoscore is a risk scoring framework that uses piecewise regressions to obtain score weights. XGBoost hyperparameters (maximum depth, learning rate, subsampling rate) were selected with 5-fold cross-validation on the training set, with  1000 maximum trees. All methods were compared on the same train, validation, and test splits, with the same covariates. Experiments were repeated 50 times on random splits to obtain performance distributions. For methods that do not make use of validation data, the training and validation sets were merged.

 Our primary objective is to assess whether transparent, human-computable risk scores can achieve predictive performance that is competitive with widely used black-box models, rather than to optimize performance at all costs. Accordingly, GBRS was evaluated using a small and fixed set of hyperparameters across all datasets, namely the number of boosting iterations, the learning rate, and the number of threshold cutoffs per variable. These parameters were chosen once and held constant to reflect realistic usage scenarios in which interpretability and reproducibility are prioritized over extensive model tuning.

In contrast, XGBoost was tuned via cross-validation over several hyperparameters to reflect its typical use as a high-performance baseline. As a result, the comparison intentionally favors XGBoost in terms of optimization flexibility. Despite this, GBRS consistently achieved performance that was close to that of XGBoost on many tasks, while producing substantially more compact and interpretable models. These results suggest that GBRS provides a favorable trade-off between predictive accuracy and transparency, supporting its use in settings where model interpretability and manual evaluability are essential.

Although risk scores are most common in binary and time-to-event settings, the same point-based, human-computable scoring paradigm is also used for continuous outcomes (e.g., severity indices, utilization scores, cost or length-of-stay estimation). We therefore evaluate GBRS across regression, classification, and survival tasks. Twelve different datasets with binary, continuous, and survival objectives were selected, 4 for each model class. We selected the open datasets Housing \cite{noauthor_housing_nodate}, Abalone \cite{warwick_nash_abalone_1994}, Diabetes \cite{noauthor_diabetes_nodate}, Cardio \cite{noauthor_cardiovascular_nodate}, Wine \cite{stefan_aeberhard_wine_1992}, Insurance \cite{noauthor_insurance_nodate}, and HELOC \cite{noauthor_home_nodate} based on their size and objective that could apply to risk scores. We also constructed four survival and one binary objectives from the UK Biobank \cite{sudlow_uk_2015} dataset for predicting future risk of heart failure, chronic kidney disease, COPD, and diabetes. The pre-processing of UK Biobank data is described in Supplementary section \ref{supp_ukb}. 

We compare methods based on their mean-squared error, AUC, and C-index performance on the held-out test set for the corresponding tasks. Results are shown in Table \ref{table_res}. On most datasets, the performance differences between methods are minimal. We generally observe that XGBoost performs significantly better on continuous objectives, and similarly well on binary and time-to-event tasks. This pattern was observed in other benchmarks \cite{dolezalova_development_2021}. We tested for differences in performance between AutoScore and GBRS. We first applied the Shapiro–Wilk test to assess whether the distribution of score differences followed a normal distribution. If normality was confirmed, we used a paired t-test; otherwise, we used the Wilcoxon signed-rank test. GBRS performed statistically better after correcting for multiple testing with the Benjamini-Hochberg procedure in all four binary classification tasks, and one survival task, while the others were non-significant.

Additionally, if scores were to be obtained from regression weights, weighted coefficients would need to be binned as a post-processing step, which could compromise the model's predictive capabilities. Although XGBoost tends to perform better on average, and is still considered a state-of-the-art method for tabular data prediction, variable importance can only be approximated post-hoc with explanation methods such as Shapley values \cite{lundberg_unified_2017}. In contrast, weights given to each variable can be directly visualized in the output of risk scores such as generated by our method. 

We show an example of a risk score for heart failure obtained in the UK Biobank in Table \ref{score_ex}, and examples of risk scores for other diseases in Supplementary Section \ref{supp_rs_examples}. In this example, we observe that smoking increases the hazard ratio for the heart failure hospitalization risk coefficient by 1.1. This corresponds to a hazard ratio of  $e^{1.1} \approx 3$, which is in line or slightly higher with hazard ratio estimates on smoking risk from other studies \cite{kamimura_cigarette_2018, aune_tobacco_2019}. This risk factor would likely be reduced by taking physical activity and other social factors into account. We can directly compare coefficients and conclude that smoking corresponds to a risk-equivalent increase in age of 10 to 20 years. We can also observe that a left ventricular ejection fraction (LVEF) of less than 50\% increases the risk of heart failure hospitalization by more than sixfold.

\begin{table}
\begin{Verbatim}[fontsize=\small, xleftmargin=0pt]

 ===================================================== 
| Age   | <50.0 | [50.0,60.0) | [60.0,70.0) | >=70.0 
|       | 0.0   | 0.3         | 0.9         | 1.6     
 ===================================================== 
| Sex   | F     | M 
|       | 0.0   | 0.4   
 ======================= 
| BMI   | <25.0 | [25.0,30.0) | >=30.0 
|       | 0.0   | 0.5         | 0.9     
 ======================================= 
| SBP   | <120.0 | [120.0,130.0) | [130.0,140.0) | >=140.0 
|       | 0.1    | 0.0           | 0.1           | 0.4      
 =========================================================== 
| LVEF  | <50.0 | >=50.0 
|       | 1.9   | 0.0         
 ===============================
| Smoking  | FALSE | TRUE 
|          | 0.0   | 1.1   
 ============================= 
\end{Verbatim}
\caption{Example of a risk score obtained with GBRS for the longitudinal risk of hospitalization of heart failure (CCSR code CIR019) in the UK Biobank. Points directly correspond to variations in log-hazard ratio. LVEF: Left Ventricular Ejection Fraction}
\label{score_ex}
\end{table}

\begin{table}[t]
\centering
\small

% --- MSE ---
\begin{tabularx}{0.8\linewidth}{lXXXX}
\toprule
MSE $\downarrow$ & Abalone & Housing & Insurance & Wine\\
\midrule
Lin.\ Reg& 2.452 $\pm$ 0.046 & 7.006 $\pm$ 0.041 & 6.019 $\pm$ 0.278 & 0.674 $\pm$ 0.021\\
XGBoost  & \textbf{2.399 $\pm$ 0.057} & \textbf{6.061 $\pm$ 0.048} & \textbf{4.554 $\pm$ 0.431} & \textbf{0.603 $\pm$ 0.022}\\
GBRS     & 2.637 $\pm$ 0.054 & 7.247 $\pm$ 0.049 & 5.970 $\pm$ 0.258 & 0.652 $\pm$ 0.023\\
\bottomrule
\end{tabularx}

\vspace{0.6em}

% --- AUC ---
\begin{tabularx}{0.8\linewidth}{lXXXX}
\toprule
AUC $\uparrow$ & UKB & Cardio & Diabetes & Heloc\\
\midrule
Log.\ Reg & 0.745 $\pm$ 0.019 & 0.782 $\pm$ 0.004 & \textbf{0.825 $\pm$ 0.024} & 0.789 $\pm$ 0.003\\
AutoScore & 0.720 $\pm$ 0.023 & 0.784 $\pm$ 0.002 & 0.801 $\pm$ 0.018 & 0.764 $\pm$ 0.039\\
XGBoost   & \textbf{0.756 $\pm$ 0.014} & \textbf{0.801 $\pm$ 0.002} & 0.820 $\pm$ 0.026 & \textbf{0.801 $\pm$ 0.004}\\
GBRS      & 0.743 $\pm$ 0.014 * & 0.786 $\pm$ 0.002 * & 0.817 $\pm$ 0.027 * & 0.776 $\pm$ 0.003 *\\
\bottomrule
\end{tabularx}

\vspace{0.6em}

% --- C-Index ---
\begin{tabularx}{0.8\linewidth}{lXXXX}
\toprule
C-Index $\uparrow$ & Heart Failure & Diabetes & CKD & COPD\\
\midrule
Cox Reg.  & 0.793 $\pm$ 0.036 & \textbf{0.713 $\pm$ 0.021} & \textbf{0.698 $\pm$ 0.038} & 0.747 $\pm$ 0.050\\
AutoScore & 0.804 $\pm$ 0.043 & 0.693 $\pm$ 0.022 & 0.673 $\pm$ 0.047 & 0.712 $\pm$ 0.053\\
XGBoost   & \textbf{0.807 $\pm$ 0.040} & 0.696 $\pm$ 0.025 & 0.696 $\pm$ 0.040 & 0.733 $\pm$ 0.043\\
GBRS      & 0.797 $\pm$ 0.043 & 0.698 $\pm$ 0.023 & 0.674 $\pm$ 0.032 & \textbf{0.749 $\pm$ 0.044} *\\
\bottomrule
\end{tabularx}

\caption{Held-out test set performance of regression methods, XGBoost, AutoScore, and GBRS. The best method for each dataset is highlighted in bold. Stars highlight cases where differences were statistically significant between AutoScore and GBRS. CKD: chronic kidney disease, COPD: chronic obstructive pulmonary disease.}
\label{table_res}
\end{table}

\begin{figure}
\includegraphics[width=\textwidth]{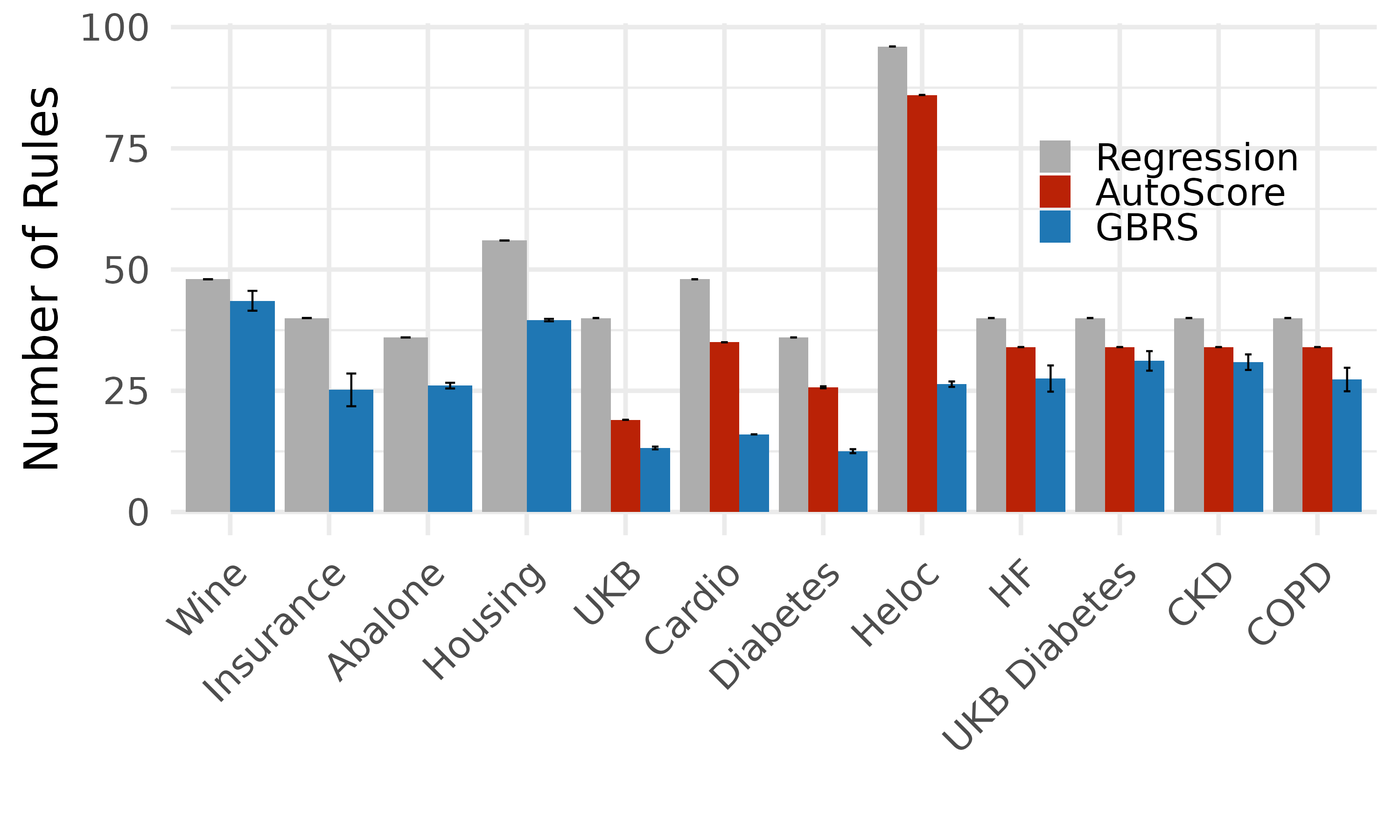}
\caption{ Size by number of rules (or stumps) of scores generated by Autoscore and GBRS on benchmarked datasets, averaged over 50 runs. We also estimate the number of rules that would be generated by a regression model from the number of input variables and thresholds. For classification tasks, the average number of rules was 41.4 for AutoScore and 17.0 for GBRS. For time-to-event tasks, 34 for AutoScore and 29.2 for GBRS. HF: heart failure, CKD: chronic kidney disease, COPD: chronic obstructive pulmonary disease.}
\label{n_params_plots}
\end{figure}

As GBRS obtains scores from a gradient boosting procedure, some variable threshold cutoffs, whether they are pre-specified by users or computed from the training set distribution, are not used in the model. In contrast, scores from regression-based techniques linearly interpolate regression weights, resulting in scores composed of a greater number of rules (or stumps). We quantified the average number of rules needed for AutoScore and GBRS scores for all binary and time-to-event tasks in Figure \ref{n_params_plots}. GBRS scores were, on average, composed of 60\% fewer rules in classification tasks and 16\% in time-to-event tasks with the same number of initial thresholds.

In practice, it is common for variable cutoffs to be pre-specified by users. For instance, BMI cutoffs should be based on commonly used weight class categories, instead of distribution characteristics. We allow for variable thresholds to be specified in GBRS as optional arguments. The sparseness of scores generated, combined with the ability to model non-linear relationships, makes GBRS a valuable tool for data analysis. In the UK-Biobank derived risk scores, we can observe and validate multiple relationships between covariates and the risk of disease. These relations could be modeled with interaction effects, but would require the user to have a prior hypothesis. For instance, an LVEF $< 50\%$ is an essential marker in heart failure \cite{shams_left_2025}, and its effect on future event risk can be observed directly from the score. Approximating the relationship between LVEF and risk as linear would not capture this clinically validated criterion.

\section{Discussion}

In this work, we present GBRS, a gradient boosting-based approach for generating risk scores. We evaluated GBRS on 16 different datasets in regression, binary classification, and time-to-event tasks. We showed that GBRS performs well on all measured objectives and outputs significantly sparser risk scores than considered alternatives. We also showed how the coefficients of GBRS scores could be readily interpreted for data analysis. We provide a C++ implementation of the risk scoring algorithm with Python and R bindings.

Unlike other risk scoring approaches, attributed points directly reflect modulations in predictions. Points represent direct changes in outcome, log odds, and hazard ratio for continuous, binary, and survival objectives, respectively. This allows the user to rescale the phenotype as needed to obtain valid and predictive scores. Due to the nonparametric nature of gradient boosting methods, our approach can capture non-linear relationships without making prior assumptions about the parameter distribution. Risk scores can overfit when the number of bins for a variable is too large with respect to the number of samples. Restricting this number acts as a natural regularization mechanism and removes the need for an additional tunable regularization coefficient.

Risk scores have several limitations that may impact their predictive capabilities. For instance, if the score is interpreted literally, we predict that a 70-year-old individual has twice the risk of heart failure compared to a 69-year-old. This is not representative of reality, and the age-risk relationship for CVD is approximately linear within small age ranges. Therefore, in some cases, it may be appropriate to linearize the weights across score thresholds. There are variables for which this would not be adequate; a LVEF $< 50\%$ is a clinically accepted disease marker, thus it is better captured with hard thresholds and should not be linearized. On some regression datasets, other methods outperform GBRS, which is consistent with approximately linear relationships and the coarse discretization implied by point-based models. GBRS is intended for settings where transparency is required and justifies this trade-off.

As risk scores cannot model variable interactions and are intended to rely on a relatively small number of input variables, their effective use requires careful variable selection and feature engineering. In addition, risk scores may lose predictive power when extrapolated far beyond the training data, particularly in settings where approximately linear relationships remain valid outside the observed sample range. Consequently, risk score models are best applied to new data that are reasonably close to the training distribution.

These limitations reflect deliberate design choices rather than deficiencies of the approach. By prioritizing transparency, stability, and human interpretability over maximal predictive flexibility, GBRS is well suited to high-stakes domains such as clinical decision-making and risk stratification, where understanding and trusting model outputs is essential. In such settings, the ability to express predictions as compact, auditable point-based scores provides practical value that complements more complex black-box models.

\section{Acknowledgements}

This work was funded by the Swiss National Science Foundation under grant CRSII5\_202276. This research has been conducted using the UK Biobank Resource under Application Number 80108.  This work uses data provided by patients and collected by the NHS as part of their care and support.

\newpage
\printbibliography
\newpage
\section{Supplementary}

\subsection{UK Biobank Preprocessing}
\label{supp_ukb}
The UK Biobank (UKB) is a large-scale and comprehensive observational study. It contains in-depth health and genetic information for 500'000 volunteer participants. Many modalities of data are available in UKB, including physical measures, questionnaire questions, multiple modalities of imaging, whole genome sequencing, and hospitalization events. 

We mapped hospitalization events represented as ICD-10 codes to Clinical Classifications Software Refined (CCSR) \cite{noauthor_clinical_nodate} v2023.1 categories. CCSR is a classification system developed by the US Agency for Healthcare Research and Quality's Healthcare Cost and Utilization Project, which aggregates ICD-10 codes into clinically meaningful categories. We selected heart failure (HF, CIR019), diabetes (END002), chronic kidney disease (CKD, GEN003), and chronic obstructive pulmonary disease (COPD, RSP008) as diseases of interest.  

Subjects with prior events for each disease were discarded and automatically censored after ten years. For the binary benchmark, we selected clinically relevant variables available at the second time point, namely age, body mass index (BMI), left ventricular ejection fraction (LVEF), and current smoking habit, which were available for 31975 subjects. We also included the left ventricular end-diastolic volume index (LVEDVI), the FEV1/FVC ratio, and the right ventricular ejection fraction (RVEF) when generating disease-specific printable scores.

\subsection{Examples of GBRS Clinical Risk Scores}
\label{supp_rs_examples}

\begin{table}[h]
\begin{Verbatim}[fontsize=\small]
CKD
 ===================================================== 
| Age | <50.0 | [50.0,60.0) | [60.0,70.0) | >=70.0 
|     | 0.0   | 0.6         | 1.6         | 2.1     
 ===================================================== 
| Sex | F     | M 
|     | 0     | 0.2   
 ======================= 
| BMI | <25.0 | [25.0,30.0) | >=30.0 
|     | 0.0   | 0.3         | 0.8     
 ======================================= 
| LVEF | <50.0 | [50.0,60.0) | >=60.0 
|      | 0.5   | 0.0         | 0.1     
 ========================================= 
| RVEF | [,0.4) | >=0.4 
|      | 0.0    | 0.3    
 ============================= 
| LVEDVI | <60.0 | [60.0,80.0) | >=80.0 
|        | 0.6   | 0.1         | 0.0     
 =========================================== 
\end{Verbatim}
\end{table}

\begin{table}
\begin{Verbatim}[fontsize=\small]
COPD
 ===================================================== 
| Age | <50.0 | [50.0,60.0) | [60.0,70.0) | >=70.0 
|     | 0.0   | 2.2         | 2.4         | 2.7     
 ===================================================== 
| Sex | F     | M  
|     | 0     | 0.5   
 ======================= 
| BMI | <25.0 | [25.0,30.0) | >=30.0 
|     | 0.0   | 0.6         | 0.7     
 ======================================= 
| LVEF | <50.0 | [50.0,60.0) | >=60.0 
|      | 1.0   | 0.0         | 0.2     
 ========================================= 
| RVEF | [,0.6) | >=0.6 
|      | 0.0    | 0.4    
 ============================= 
| LVEDVI | [,60.0) | >=60.0 
|        | 0.3     | 0.0     
 =============================== 
| Smoking | FALSE | TRUE 
|         | 0     | 2.2   
 ============================= 
\end{Verbatim}
\end{table}

\begin{table}
\begin{Verbatim}[fontsize=\small]
Diabetes
 ===================================================== 
| Age | <50.0 | [50.0,60.0) | [60.0,70.0) | >=70.0 
|     | 0.0   | 0.2         | 0.3         | 0.7     
 ===================================================== 
| Sex | F     | M  
|     | 0     | 0.7   
 ======================= 
| BMI | <25.0 | [25.0,30.0) | >=30.0 
|     | 0.0   | 1.0         | 2.0     
 ======================================= 
| LVEF | <50.0 | [50.0,60.0) | >=60.0 
|      | 0.5   | 0.0         | 0.4     
 ========================================= 
| RVEF | <0.4 | [0.4,0.6) | >=0.6 
|      | 0.6  | 0.0       | 0.4    
 ======================================= 
| LVEDVI | <60.0 | [60.0,80.0) | >=80.0 
|        | 0.9   | 0.4         | 0.0     
 =========================================== 
\end{Verbatim}
\end{table}

\end{document}